\documentclass[english]{svproc}

\usepackage[latin1]{inputenc}
\usepackage{amsmath}
\usepackage{amssymb}
\usepackage{mathrsfs}
\usepackage{esint}
\usepackage{graphicx}
\usepackage{array}
\usepackage{rotfloat}

\usepackage{array}   %stellt den Befehl \newcolumntype bereit
\newcolumntype{C}[1]{>{\centering\arraybackslash}p{#1}} 
\usepackage{booktabs}
\newcolumntype{L}[1]{>{\raggedright\arraybackslash}p{#1}}
\newcolumntype{R}[1]{>{\raggedleft\arraybackslash}p{#1}}
\usepackage{multirow}
\usepackage{comment}
\usepackage{rotating}

\makeatletter

\newcommand{\noun}[1]{\textsc{#1}}
\providecommand{\tabularnewline}{\\}

\usepackage{enumitem}%alpanumeric counting in \enumerate
\usepackage{bibspacing}

\usepackage{babel}
\begin{document}

	\title{Activation Functions for Generalized Learning Vector Quantization - A Performance Comparison \vspace{1cm}\\
	\emph{--Draft version of a regarding paper submitted to the WSOM+2019 conference to be held in Barcelona, June 2019 --}}
	\titlerunning{Activation Functions for GLVQ}
	
	\author{T.~Villmann\inst{1}\thanks{\emph{corresponding author}, email: thomas.villmann@hs-mittweida.de} \and J.~Ravichandran\inst{1} \and A.~Villmann\inst{1,2} \and \\ D.~Nebel \and M.~Kaden\inst{1} }
	\authorrunning{Thomas Villmann et al.}
	\tocauthor{Thomas Villmann, John Ravichandran, Andrea Villmann, David Nebel, Marika Kaden}
	
	\institute{
		Saxony Institute for Computational Intelligence and Machine Learning,\\
		University of Applied Sciences Mittweida, Germany \\ 
		\texttt{https://www.institute.hs-mittweida.de/webs/sicim.html} \vspace{0.2cm}
			%http://users/\homedir iekeland/web/welcome.html
		\and
		Schulzentrum Döbeln-Mittweida, Germany
	}

	\maketitle

\begin{abstract}
An appropriate choice of the activation function plays an important role in the performance of (deep) multilayer perceptrons (MLP) for classification and regression learning.  Prototype-based classification learning methods like (generalized) learning vector quantization (GLVQ) are powerful alternatives. These models also deal with activation functions but here they are applied to the so-called classifier function instead. In this paper we investigate successful candidates of activation functions known for MLPs for application in GLVQ and their influence on the performance. 
\end{abstract}

\section{Introduction}

Prototype-based classification learning like learning vector quantization
(LVQ) was introduced by \noun{T. Kohonen} in \cite{kohonen88j} and belongs
to robust and stable classification models in machine learning \cite{Villmann2018g}.
One of the most prominent variants is generalized learning vector
quantization (GLVQ, \cite{sato96a}). The GLVQ cost function to be
minimized by stochastic gradient descent learning (SGDL) is an approximation
of the overall classification error. Further, GLVQ belongs
to the family of margin optimizers for classification learning, because
GLVQ maximizes the hypothesis margin \cite{Crammer2002a}.

Beside the usually taken geometric perspective for GLVQ interpretation, the neural network perspective of LVQ gains more and more attraction,
because it allows to combine LVQ models with techniques of deep learning
\cite{deVries:DeepLVQ:ESANN2016,Villmann2017h}. In this perspective
the LVQ prototypes, as elements of the Euclidean space, are interpreted as weight vectors of linear perceptrons,
such that the maximum perceptron excitation by a data vector corresponds to minimum Euclidean distance according to the nearest prototype principle realized in vector quantization approaches \cite{kohonen95a}. GLVQ performance crucially depends on the activation function for the so-called classifier function of the GLVQ costs, usually chosen as a parametrized sigmoid function \cite{Villmann2015d}.

A hot topic in deep network design is the search for appropriate perceptron activation functions improving standard sigmoid or ReLU \cite{GoodfellowEtAl:DeepLearning:BookMITPress2016,RamachandranEtAl:SwishSelfGatedActivationFunction:arXiv2018}. Yet, recent studies show that ReLU-units can be further improved using more sophisticated activation function like \emph{swish}, \emph{soft+}
and others \cite{RamachandranEtAl:SearchingForActivationFunctions:arXiv2018,EgerEtAl:TimeToSwish:ProcEMNLP,ChiengEtAl:FlattenTSwisshThresholdedSwish:JournAdvIntellSystems2018}.

For GLVQ, to our best knowledge, only linear and sigmoid activation
functions were considered as activation functions for the (GLVQ-) classifier function regarding their classification behavior whereas
convergence behavior was not in the focus so far. 

Therefore, the
aim of this contribution is to investigate several prominent state-of-the-art
activation functions of (deep) MLPs regarding their convergence behavior
and resulting final classification performance when applied in GLVQ.

\section{Generalized Learning Vector Quantization - from a Multilayer Network Perspective}

We start considering GLVQ from the geometric perspective.

In GLVQ, a set $W=\left\{ \mathbf{w}_{1},\mathbf{w}_{2},...,\mathbf{w}_{N}\right\} $
of prototypes $\mathbf{w}_{k}\in\mathbb{R}^{p}$ is assumed as well
as (training) data $\mathbf{x}\in X\subseteq\mathbb{R}^{n}$  equipped with
class labels $c\left(\mathbf{x}\right)$. Each prototype is uniquely
responsible for a certain class $c_{k}=c\left(\mathbf{w}_{k}\right)$.
The data are projected by means of a projection $\pi:\mathbb{R}^{n}\rightarrow\mathbb{R}^{p}$
into the prototype space $\mathbb{R}^{p}$ also denoted as projection
space in this context. The vector quantization mapping $\mathbf{x}\mapsto\kappa\left(\mathbf{x}\right)$
takes place as a winner-takes-all (WTA) rule according to
\begin{equation}
\kappa\left(\mathbf{x},W\right)=\textrm{argmin}_{k:\mathbf{w}_{k}\in W}\left\{ d\left(\pi\left(\mathbf{x}\right),\mathbf{w}_{k}\right)|k=1,2,...N\right\} .\label{eq:wta}
\end{equation}
realizing the nearest prototype principle with respect to a predefined
dissimilarity measure $d$. The value $\kappa\left(\mathbf{x},W\right)$
is called the index of the best matching (winner) prototype with respect
to the set $W$. An unknown data vector $\mathbf{u}$ is assigned
to the class $c_{\kappa\left(\mathbf{u}\right)}=c\left(\mathbf{w}_{\kappa\left(\mathbf{u}\right)}\right)$.

Usually, the (squared) Euclidean distance is used, which can be written as 
\begin{equation}
d_{\pi}\left(\mathbf{x},\mathbf{w}_{k}\right)=-2\left\langle \pi\left(\mathbf{x}\right),\mathbf{w}_{k}\right\rangle _{E}-b_{k}\left(\mathbf{x}\right)\label{eq:Euclidean projection distance}
\end{equation}
where $\left\langle \mathbf{z},\mathbf{w}_{k}\right\rangle _{E}$
denotes the Euclidean inner product and 
\begin{equation}
b_{k}\left(\mathbf{x}\right)=\left\langle \pi\left(\mathbf{x}\right),\pi\left(\mathbf{x}\right)\right\rangle +\left\langle \mathbf{w}_{k},\mathbf{w}_{k}\right\rangle _{E}
\end{equation} collects the squared norm values of $\mathbf{x}$ and $\mathbf{w}$. 
Let $\mathbf{w}^{+}\in W^{+}$ and $\mathbf{w}^{-}\in W^{-}$ be the
best matching prototypes according to $W^{+}\left(\mathbf{x}\right)=\left\{ \mathbf{w}_{k}|\mathbf{w}_{k}\in W\wedge c_{k}=c\left(\mathbf{x}\right)\right\} \subset W$
and $W^{-}\left(\mathbf{x}\right)=W\setminus W^{+}$, respectively.
The local loss in GLVQ is defined as 
\begin{equation}
l\left(\mathbf{x},W,\gamma\right)=f\left(\frac{d_{\pi}^{+}\left(\mathbf{x}\right)-d_{\pi}^{-}\left(\mathbf{x}\right)}{\eta\left(\mathbf{x}\right)}-\gamma\right)\label{eq:local cost GLVQ}
\end{equation}
with $d_{\pi}^{\pm}\left(\mathbf{x}\right)=d_{\pi}\left(\mathbf{x},\mathbf{w}^{\pm}\right)$,
the \emph{GLVQ-activation function}$f$ is a monotonically increasing
and differentiable function, which is denoted as \emph{activation function} for the classifier function 
\begin{equation}
\mu\left(\mathbf{x},\gamma\right)=\frac{d_{\pi}^{+}\left(\mathbf{x}\right)-d_{\pi}^{-}\left(\mathbf{x}\right)}{\eta\left(\mathbf{x}\right)}-\gamma \;.
\end{equation} 
The quantity $\eta\left(\mathbf{x}\right)=d_{\pi}^{+}\left(\mathbf{x}\right)+d_{\pi}^{-}\left(\mathbf{x}\right)$
is the local normalization, and $\gamma\in\mathbb{R}$ is a shifting
variable frequently set to zero. The difference quantity $h\left(\mathbf{x}\right)=\frac{1}{2}\left|d_{\pi}^{-}\left(\mathbf{x}\right)-d_{\pi}^{+}\left(\mathbf{x}\right)\right|$
is denoted as (local) \emph{hypothesis margin} \cite{Crammer2002a},
which is related to the \emph{hypothesis margin vector} 
\begin{equation}
\mathbf{h}\left(\mathbf{x}\right)=\mathbf{w}^{-}-\mathbf{w}^{+}\label{eq:hypothesis margin vector}
\end{equation}
via the triangle $\triangle\left(\mathbf{x},\mathbf{w}^{+},\mathbf{w}^{-}\right)$.
The classifier function yields negative values only
for correctly classified training samples.

The cost function to be minimized by \emph{stochastic gradient descent
learning} (SGDL) becomes
\begin{equation}
E_{GLVQ}\left(X,W\right)=\sum_{\mathbf{x}\in X}l\left(\mathbf{x},W\right)\label{eq:GLVQ cost function}
\end{equation}
as explained in \cite{sato96a}. 
Doing so, the stochastic gradient of the cost function involves the  derivative $\frac{\partial f\left(\mu\left(\mathbf{x},\gamma\right)\right)}{\partial\mu\left(\mathbf{x},\gamma\right)}$
of the activation function.

The choice $\pi\left(\mathbf{x}\right)=\mathbf{x}$
yields the standard GLVQ whereas for the linear mapping $\pi_{\boldsymbol{\Omega}}\left(\mathbf{x}\right)=\boldsymbol{\Omega}\mathbf{x}$
the matrix variant GMLVQ is obtained, which reduces to relevance GLVQ
(GRLVQ) for a diagonal matrix $\boldsymbol{\Omega}$ \cite{Schneider2009_MatrixLearning,Villmann2002d}. If $\pi\left(\mathbf{x}\right)$ is realized by a deep network, DeepGLVQ is resulted \cite{Villmann2017h,deVries:DeepLVQ:ESANN2016}.

In the following we reconsider GLVQ taking the neural network perspective.
For this purpose, remark that the squared Euclidean distance \eqref{eq:Euclidean projection distance} can be seen as a linear perceptron with weight vector $\mathbf{w}_{k}$
and the bias $b_{k}\left(\mathbf{x}\right)$
regarding the projected input vector $\pi\left(\mathbf{x}\right)\in\mathbb{R}^{p}$
\cite{Villmann2018j}. As shown in \cite{Villmann2019c}, we get for the local costs (\ref{eq:local cost GLVQ}) 
\begin{eqnarray*}
l\left(\mathbf{x},W\right) & = & f\left(\left\langle \hat{\pi}\left(\mathbf{x}\right),\mathbf{h}\left(\mathbf{x}\right)\right\rangle _{E}+B^{\pm}\left(\mathbf{x},\gamma\right)\right)\\
 & = & \Pi_{f}\left(\mathbf{x},W\right)
\end{eqnarray*}
with the scaled data mapping $\hat{\pi}\left(\mathbf{x}\right)=\frac{2\pi\left(\mathbf{x}\right)}{\eta\left(\mathbf{x}\right)}$
and hypothesis margin vector $\mathbf{h}\left(\mathbf{x}\right)$. Thus,
the local loss can be seen as a linear perceptron with hypothesis
margin vector $\mathbf{h}\left(\mathbf{x}\right)$ as the weight vector
and a parameterized bias $B^{\pm}\left(\mathbf{x},\gamma\right)$ for
the projected data $\hat{\pi}\left(\mathbf{x}\right)$. In this sense,
the GLVQ activation function can be interpreted as an activation function for the special
\emph{GLVQ-perceptron} $\Pi_{f}\left(\mathbf{x},W\right)$. Note that the GLVQ-perceptron\emph{
$\Pi_{f}\left(\mathbf{x},W\right)$} delivers maximum local costs
for maximum excitation, such that GLVQ-classification-learning relates
to minimum excitation learning for GLVQ-perceptrons\emph{ $\Pi_{f}\left(\mathbf{x},W\right)$}.
In GLVQ, standard choices for the activation function are identity
$\textrm{id}\left(x\right)=x$ and the sigmoid $\textrm{sgd}\left(x,\beta\right)$
with $\beta=1$, see Tab.$\,$\ref{table:Activation functions and derivatives}.

\subsection{Activation Function for MLP and GLVQ-MLN}

As we have explained in the previous subsection, the local loss in
GLVQ can be described as a particular perceptron structure. Hence,
the consideration of the respective activation function becomes inevitable.
Many considerations for (deep) MLPs have shown that the appropriate
choice of activation functions is essential for convergence behavior
and final network performance \cite{ChiengEtAl:FlattenTSwisshThresholdedSwish:JournAdvIntellSystems2018}.
Originally, sigmoid functions like \emph{tangens-hyperbolicus} or
standard sigmoid function $\textrm{sgd}\left(x,\beta\right)$ with
$\beta=1$ (see Tab.$\,$(\ref{table:Activation functions and derivatives}))
were preferred to ensure non-linearity and differentiability together
with easy analytical computation of derivatives. Later, \emph{Rectified
linear Units} (ReLU)
$\textrm{ReLU}\left(x\right)=\max\left(0,x\right)$ became popular
due to its performance and computational simplicity \cite{GoodfellowEtAl:DeepLearning:BookMITPress2016}.
Recently, a systematic study of activation functions was proposed 
proposed \cite{RamachandranEtAl:SearchingForActivationFunctions:arXiv2018}.  It turns
out that the \emph{swish}-function $\textrm{swish}\left(x,\beta\right)=x\cdot\textrm{sgd}\left(x,\beta\right)$
introduced in \cite{ElfwingEtAl:SwishforNeuralNetworksFunctionApproximation:NeuralNetworks2018}
is, in average, the  most successful although not always the best choice. It can be seen as an intermediate
between $\textrm{ReLU}\left(x\right)$ and the scaled identity $\textrm{id}\left(x\right)$
according to \emph{functional} limits
\begin{equation}
\textrm{swish}\left(x,\beta\right)\underset{\beta\rightarrow\infty}{\longrightarrow}\textrm{ReLU}\left(x\right)
\textrm{ and }
\textrm{swish}\left(x,\beta\right)\underset{\beta\searrow0}{\longrightarrow}\textrm{id}\left(x\right)\label{eq:Swish ReLU Id Approximation}
\end{equation}
respectively. Yet, other activation functions like $m\left(x,\beta\right)=\max\left(x,\textrm{sgd}\left(x,\beta\right)\right)$
also perform very well for deep MLP as outlined in \cite{RamachandranEtAl:SearchingForActivationFunctions:arXiv2018}.
Further, the choice of the activation also affects the classification
robustness \cite{ZhangEtal:EfficientNeuralNetworkRobustnessCertificationWithGeneralActivationFunctions:NIPS2018_7742}.
A collection of promising activation functions together with their derivatives\footnote{The derivative of the maximum function $m\left(x,\beta\right)$ could
	be approximated using the quasi-max function $\mathcal{Q}_{\alpha}\left(x,\beta\right)=\frac{1}{\alpha}\log\left(e^{\alpha x}+e^{\alpha\cdot\textrm{sgd}\left(x,\beta\right)}\right)$
	proposed by \noun{J.D. Cook} \cite{CookSoftmax2011} with $\alpha\gg0$.
	The respective consistent derivative approximation is 
	\begin{equation}
	\frac{d\,m\left(x,\beta\right)}{dx}\approx\frac{\left(\exp\left(\alpha x\right)+\frac{d\textrm{sgd}\left(x,\beta\right)}{dx}\cdot\exp\left(\alpha\cdot\textrm{sgd}\left(x,\beta\right)\right)\right)}{\left(\exp\left(\alpha x\right)+\exp\left(\alpha\cdot\textrm{sgd}\left(x,\beta\right)\right)\right)}\label{eq: approx derivative max function}
	\end{equation}
	as provided in \cite{Villmann2013m}. Analogously, the quasi-max
	approximation $m_{\tau}\left(x,\beta\right)\approx\frac{1}{\alpha}\log\left(\exp\left(\alpha x\right)+\exp\left(\alpha\cdot\tau\left(x,\beta\right)\right)\right)$
	is valid with
	\begin{equation}
	\frac{d\,m_{\tau}\left(x,\beta\right)}{dx}\approx\frac{\left(\exp\left(\alpha x\right)+\frac{d\,\tau\left(x,\beta\right)}{dx}\cdot\exp\left(\alpha\cdot\tau\left(x,\beta\right)\right)\right)}{\left(\exp\left(\alpha x\right)+\exp\left(\alpha\cdot\tau\left(x,\beta\right)\right)\right)}\label{eq: approx derivative max tau function}
	\end{equation}
	as the derivative approximation.} 
is given in Tab.$\,$\ref{table:Activation functions and derivatives}. 
Note that for $\beta=0$ the \emph{Leaky ReLU} $\textrm{LReLU}\left(x,\beta\right)$
introduced in \cite{MaasEtAl:RectifierNonlinearitiesImproveNeuralNetworkAcousticModels:ICML2013}
simply becomes $\textrm{ReLU}\left(x\right)$, whereas $\textrm{swish}_{\tau}\left(x,\beta\right)$
and $m_{\tau}\left(x,\beta\right)$ are variants of $\textrm{swish}\left(x,\beta\right)$
and $m\left(x,\beta\right)$ replacing the sigmoid $\textrm{sgd}\left(x,\beta\right)$
by the tangens-hyperbolicus function $\tau\left(x,\beta\right)$.

\begin{sidewaystable}
	\begin{tabular}{|c|c|}
		\hline 
		activation function & derivative\tabularnewline
		\hline 
		\hline 
		$\textrm{sgd}\left(x,\beta\right)=\frac{1}{1+\exp\left(-\beta\cdot x\right)}$ & $\frac{d\textrm{sgd}\left(x,\beta\right)}{dx}=\beta\cdot\textrm{sgd}\left(x,\beta\right)\cdot\left(1-\textrm{sgd}\left(x,\beta\right)\right)$\tabularnewline
		\hline 
		$\tau\left(x,\beta\right)=\tanh\left(x\cdot\beta\right)+1$ & $\frac{d\tau\left(x,\beta\right)}{dx}=\beta\cdot\left(1-\left(\tanh\left(x,\beta\right)\right)^{2}\right)$\tabularnewline
		\hline 
		$\textrm{swish}\left(x,\beta\right)=x\cdot\textrm{sgd}\left(x,\beta\right)$ & $\frac{d\textrm{swish}\left(x,\beta\right)}{dx}=\beta\cdot\textrm{swish}\left(x,\beta\right)+\textrm{sgd}\left(x,\beta\right)\cdot\left(1-\beta\cdot\textrm{swish}\left(x,\beta\right)\right)$\tabularnewline
		\hline 
		$\textrm{swish}_{\tau}\left(x,\beta\right)=x\cdot\tau\left(x,\beta\right)$ & $\frac{d\textrm{swish}_{\tau}\left(x,\beta\right)}{dx}=x\cdot\beta+\tau\left(x,\beta\right)\cdot\left(1-\beta\cdot\textrm{swish}_{\tau}\left(x,\beta\right)\right)$\tabularnewline
		\hline 
		$\textrm{LReLU}\left(x,\beta\right)=\max\left(0,\beta\cdot x\right)$ & $\frac{\partial\textrm{LReLU}\left(x\right)}{dx}=\begin{cases}
		0 & x<0\\
		\beta & x>0
		\end{cases}$ or $\frac{\partial\textrm{LReLU}\left(x\right)}{dx}=\beta\cdot\frac{\partial\textrm{ReLU}\left(x\right)}{dx}\approx\beta\cdot\frac{d\textrm{swish}\left(x,\alpha\right)}{dx}$
		for $\alpha\gg0$ acc. \eqref{eq:Swish ReLU Id Approximation}\tabularnewline
		\hline 
		$m\left(x,\beta\right)=\max\left(x,\textrm{sgd}\left(x,\beta\right)\right)$ & $\frac{\partial\,m\left(x,\beta\right)}{dx}=\begin{cases}
		1 & x>\textrm{sgd}\left(x,\beta\right)\\
		\frac{d\textrm{sgd}\left(x,\beta\right)}{dx} & x<\textrm{sgd}\left(x,\beta\right)
		\end{cases}$ or $\frac{d\,m\left(x,\beta\right)}{dx}\approx\frac{\left(\exp\left(\alpha x\right)+\frac{d\textrm{sgd}\left(x,\beta\right)}{dx}\cdot\exp\left(\alpha\cdot\textrm{sgd}\left(x,\beta\right)\right)\right)}{\left(\exp\left(\alpha x\right)+\exp\left(\alpha\cdot\textrm{sgd}\left(x,\beta\right)\right)\right)}$
		for $\alpha\gg0$ acc. to (\ref{eq: approx derivative max function})\tabularnewline
		\hline 
		$m_{\tau}\left(x,\beta\right)=\max\left(x,\tau\left(x,\beta\right)\right)$ & $\frac{\partial\,m_{\tau}\left(x,\beta\right)}{dx}=\begin{cases}
		1 & x>\tau\left(x,\beta\right)\\
		\frac{d\,\tau\left(x,\beta\right)}{dx} & x<\tau\left(x,\beta\right)
		\end{cases}$ or $\frac{d\,m_{\tau}\left(x,\beta\right)}{dx}\approx\frac{\left(\exp\left(\alpha x\right)+\frac{d\,\tau\left(x,\beta\right)}{dx}\cdot\exp\left(\alpha\cdot\tau\left(x,\beta\right)\right)\right)}{\left(\exp\left(\alpha x\right)+\exp\left(\alpha\cdot\tau\left(x,\beta\right)\right)\right)}$
		for $\alpha\gg0$ acc. to (\ref{eq: approx derivative max tau function})\tabularnewline
		\hline 
		$\textrm{cosxx\ensuremath{\left(\beta,x\right)}}=\beta\cdot x-\cos\left(x\right)$ & $\frac{\partial\textrm{cosxx\ensuremath{\left(\beta,x\right)}}}{dx}=\beta+\sin\left(x\right)$\tabularnewline
		\hline 
		$\textrm{soft+}\left(\beta,x\right)=\log\left(1+\exp\left(\beta\cdot x\right)\right)$ & $\frac{\partial\textrm{soft+\ensuremath{\left(\beta,x\right)}}}{dx}=\frac{\beta\cdot\exp\left(\beta\cdot x\right)}{1+\exp\left(\beta\cdot x\right)}$\tabularnewline
		\hline 
	\end{tabular}{\caption{Successful Activation functions for MLP according to \cite{RamachandranEtAl:SearchingForActivationFunctions:arXiv2018}
			together with their derivatives.}
		\label{table:Activation functions and derivatives}}
\end{sidewaystable}

The neural network perspective of GLVQ and, particularly, the GLVQ-perceptrons \emph{ $\Pi_{f}\left(\mathbf{x},W\right)$} interpretation motivates to consider the impact of the activation functions regarding the GLVQ performance.

\section{Numerical Results}
We performed numerical investigations regarding the performance of the activation functions from Tab.$\,$\ref{table:Activation functions and derivatives} for four widely used standard data sets. 
These are
\begin{itemize}
	\item the \emph{Tecator  Data Set} comprising $215$ spectra measured for several meat probes. The spectral range is $850$ - $1050$ $nm$ with $D=100$ spectral bands. The data set is labeled according to the fat content (high/low) The data set is provided as a training set ($N_{V_{train}}=172$) and a test set ($N_{V_{test}}=43$) \cite{KrierFeatureSelectionICA2008}.\footnote{Tecator data set is available at \emph{StaLib}: http://lib.stat.cmu.edu/datasets/tecator.}
	\item the \emph{Indian Pine Data Set},which is a spectral data set from remote sensing.\footnote{The data set can be found at www.ehu.es/ccwintco/uploads/2/22/Indian\_pines.mat} It was generated by an AVIRIS sensor capturing an area corresponding to $145\times145$ pixels
	in the Indian Pine test site in the northwest of Indiana \cite{Landgrebe2003a}. The spectrometer operates in the visible and mid-infrared wavelength range ($0.4-2.4\mathrm{\mu m}$)
	with $D=220$ equidistant bands. The area includes $16$ different
	kinds of forest or other natural perennial vegetation and non-agricultural
	sectors, which are also denoted as background. These background pixels are removed from the data set as usual. Additionally, we remove $20$ wavelengths, mainly affected by water content (around
	$1.33\mathrm{\mu m}$ and $1.75\mathrm{\mu m}$. Finally, all spectral vectors
	were normalized according to the $l_{2}$-norm. This overall preprocessing
	is usually applied to this data set \cite{Landgrebe2003a}. Data classes with less than $100$ samples were removed yielding a $12$-class-problem.
	\item the \emph{Wisconsin-Breast-Cancer-data (WBCD)} and the \emph{Indian diabetes data set
	(PIMA)} contain $562$ and $768$ data vectors with
	$32$ and $8$ data dimensions, respectively, and each divided into two classes (healthy/ill). A detailed description can be found in \cite{UCIRepository}.     
\end{itemize}

\subsection{Results}

The results reported here were obtained for GLVQ with only one prototype per class.

For each data set and each activation function from Tab.$\,$\ref{table:Activation functions and derivatives} we performed $100$ runs for several parameter configurations $\beta$ by a grid search where the averaged accuracy was the evaluation criterion. The learning rate as well as a maximum number of $10000$ epochs per training experiment were maintained uniformly for all experiments. We report the results regarding the best parameter configurations. 

The detailed results can be found in \cite{Villmann2019c}. Here we give the results averaged over all four data sets. For this purpose we take $ReLU$-accuracies as references and calculate the respective ratios. Ratios greater than one indicate better accuracies, whereas lower ratios refer to worse results. The ratios averaged over all data sets together with their standard deviations are depicted in Tab.$\;$\ref{tab:results}. Additionally, we give the averaged ratios and standard deviations for convergence performance. The convergence performance is measured considering the number of training epochs until the averaged gradient becomes approximately zero. Thus, ratios lower than one refer to higher convergence rate than $ReLU$ whereas greater values indicate slower convergence.

\begin{table}
	\footnotesize
	\centering
	\begin{tabular}{|c||c|c||c|c|}
		\hline 
		activation function & av. accuracy ratio & st. dev & av. convergence ratio & st. dev.\tabularnewline
		\hline 
		\hline 
		$\textrm{ReLU}\left(x\right)$ & $1$ & $0$ & $1$ & $0$\tabularnewline
		\hline 
		$\textrm{sgd}\left(x,\beta\right)$ & $\mathbf{1.058588283}$ & $0.065044863$ & $3.292611899$ & $4.382405967$\tabularnewline
		\hline 
		$\textrm{sgd}\left(x,\beta=1\right)$ & $0.855523844$ & $0.166406655$ & $0.177755527$ & $0.089233988$\tabularnewline
		\hline 
		$\tau\left(x,\beta\right)$ & $0.896820407$ & $0.150230939$ & $0.221782303$ & $0.172070555$\tabularnewline
		\hline 
		$\textrm{swish}\left(x,\beta\right)$ & $\mathbf{1.05793573}$ & $0.058660376$ & $\mathbf{0.553609888}$ & $0.275164898$\tabularnewline
		\hline 
		$\textrm{swish}_{\tau}\left(x,\beta\right)$ & $0.936386328$ & $0.160171911$ & $0.184983626$ & $0.1057056$\tabularnewline
		\hline 
		$\textrm{LReLU}\left(x,\beta\right)$ & $1.028269063$ & $0.089099257$ & $1.023108516$ & $0.928273831$\tabularnewline
		\hline 
		$m\left(x,\beta\right)$ & $\mathbf{1.057855692}$ & $0.069498943$ & $4.519976424$ & $5.838384012$\tabularnewline
		\hline 
		$m_{\tau}\left(x,\beta\right)$ & $0.898402574$ & $0.153118782$ & $0.392754746$ & $0.385784308$\tabularnewline
		\hline 
		$\textrm{cosxx\ensuremath{\left(\beta,x\right)}}$ & $0.990139782$ & $0.10838342$ & $0.375625059$ & $0.179953065$\tabularnewline
		\hline 
		$\textrm{soft+}\left(\beta,x\right)$ & $\mathbf{1.078091745}$ & $0.058366659$ & $4.595046404$ & $3.698125022$\tabularnewline
		\hline 
		$\textrm{id}\left(x\right)$ & $0.85092197$ & $0.16492968$ & $0.141372141$ & $0.109199298$\tabularnewline
		\hline 
	\end{tabular}\vspace{0.2cm}
	\caption{Results for the activation functions compared to $ReLu$. The best accuracy results are in bold font. Among them, the best respective convergence ratio is also depicted in bold font. For further explanations see text.}
	\label{tab:results}
\end{table}

From this experiments we can conclude that $soft+$, $sgd$ and $swish$ achieve best results with almost similar accuracies for an appropriate parameter choice $\beta$, all improving standard $ReLU$.\footnote{Yet, also the maximum function $m\left(x,\beta\right)$ achieves high accuracies. However, these results are obtained for values $\beta\gg1$. In this case,  $m\left(x,\beta\right)$ behaves like $\textrm{sgd}\left(x,\beta\right)$. Therefore, it is not mentioned explicitly in the list of best functions.}  Among them, $swish$ clearly outperforms the others regarding the convergence performance. 
Further, both standard activation functions for GLVQ, $id$ and $sgd$ with $\beta=1$, are significantly weaker than $ReLU$ and, hence, also weaker than the leading activation functions. Hence they should be avoided.

\section{Conclusions}
In this paper we studied the influence of several MLP activation function candidates regarding their performance influence for GLVQ. Motivation for this investigation is the fact that the classifier function of GLVQ can be described as a generalized perceptron and, hence, the GLVQ activation function plays the role of a perceptron activation function.
The numerical experiments have shown that $soft+$, $sgd$ and $swish$ achieve the best accuracy performance better than $ReLU$ as it is also frequently the case for (deep) MLP networks \cite{RamachandranEtAl:SearchingForActivationFunctions:arXiv2018}. Yet, regarding the convergence speed $swish$ has to be highly favored. Moreover, the standard activation functions of GLVQ are clearly outperformed.

Summarizing these experiments we suggest to switch over from $id$ and $sgd$ (with $\beta=1$) to $swish$ for GLVQ activation.

\bibliographystyle{unsrt}
\addcontentsline{toc}{section}{\refname}\bibliography{C:/Users/Villy/Documents/bibtex/pub,C:/Users/Villy/Documents/bibtex/DeepLearningELM,C:/Users/Villy/Documents/bibtex/references,C:/Users/Villy/Documents/bibtex/neuro,C:/Users/Villy/Documents/bibtex/information,C:/Users/Villy/Documents/bibtex/cluster,C:/Users/Villy/Documents/bibtex/pattern,C:/Users/Villy/Documents/bibtex/zeitreih,C:/Users/Villy/Documents/bibtex/math,C:/Users/Villy/Documents/bibtex/statistics,C:/Users/Villy/Documents/bibtex/GenerativeAdversarialNetworks,C:/Users/Villy/Documents/bibtex/datasets,C:/Users/Villy/Documents/bibtex/remote_sensing}

\begin{thebibliography}{10}

\bibitem{kohonen88j}
Teuvo Kohonen.
\newblock Learning {V}ector {Q}uantization.
\newblock {\em Neural Networks}, 1(Supplement 1):303, 1988.

\bibitem{Villmann2018g}
T.~Villmann, S.~Saralajew, A.~Villmann, and M.~Kaden.
\newblock Learning vector quantization methods for interpretable classification
  learning and multilayer networks.
\newblock In C.~Sabourin, J.J. Merelo, A.L. Barranco, K.~Madani, and
  K.~Warwick, editors, {\em Proceedings of the 10th International Joint
  Conference on Computational Intelligence (IJCCI), Sevilla}, pages 15--21,
  Lissabon, Portugal, 2018. SCITEPRESS - Science and Technology Publications,
  Lda.
\newblock ISBN: 978-989-758-327-8?

\bibitem{sato96a}
A.~Sato and K.~Yamada.
\newblock Generalized learning vector quantization.
\newblock In D.~S. Touretzky, M.~C. Mozer, and M.~E. Hasselmo, editors, {\em
  Advances in Neural Information Processing Systems 8. Proceedings of the 1995
  Conference}, pages 423--9. MIT Press, Cambridge, MA, USA, 1996.

\bibitem{Crammer2002a}
K.~Crammer, R.~Gilad-Bachrach, A.~Navot, and A.Tishby.
\newblock Margin analysis of the {LVQ} algorithm.
\newblock In S.~Becker, S.~Thrun, and K.~Obermayer, editors, {\em Advances in
  Neural Information Processing (Proc. NIPS 2002)}, volume~15, pages 462--469,
  Cambridge, MA, 2003. MIT Press.

\bibitem{deVries:DeepLVQ:ESANN2016}
H.~deVries, R.~Memisevic, and A.~Courville.
\newblock Deep learning vector quantization.
\newblock In M.~Verleysen, editor, {\em Proceedings of the European Symposium
  on Artificial Neural Networks, Computational Intelligence and Machine
  Learning {(ESANN'2016)}}, pages 503--508, Louvain-La-Neuve, Belgium, 2016.
  {i6doc.com}.

\bibitem{Villmann2017h}
T.~Villmann, M.~Biehl, A.~Villmann, and S.~Saralajew.
\newblock Fusion of deep learning architectures, multilayer feedforward
  networks and learning vector quantizers for deep classification learning.
\newblock In {\em Proceedings of the 12th Workshop on Self-Organizing Maps and
  Learning Vector Quantization (WSOM2017+)}, pages 248--255. IEEE Press, 2017.

\bibitem{kohonen95a}
Teuvo Kohonen.
\newblock {\em {Self-Organizing Maps}}, volume~30 of {\em Springer Series in
  Information Sciences}.
\newblock Springer, Berlin, Heidelberg, 1995.
\newblock (Second Extended Edition 1997).

\bibitem{Villmann2015d}
M.~Kaden, M.~Riedel, W.~Hermann, and T.~Villmann.
\newblock Border-sensitive learning in generalized learning vector
  quantization: an alternative to support vector machines.
\newblock {\em Soft Computing}, 19(9):2423--2434, 2015.

\bibitem{GoodfellowEtAl:DeepLearning:BookMITPress2016}
I.~Goodfellow, Y.~Bengio, and A.~Courville.
\newblock {\em Deep Learning}.
\newblock MIT Press, 2016.

\bibitem{RamachandranEtAl:SwishSelfGatedActivationFunction:arXiv2018}
P.~Ramachandran, B.~Zoph, and Q.V. Le.
\newblock Swish: {A} self-gated activation function.
\newblock Technical Report arXiv:1710.05941v2, Google Brain, 2018.

\bibitem{RamachandranEtAl:SearchingForActivationFunctions:arXiv2018}
P.~Ramachandran, B.~Zoph, and Q.V. Le.
\newblock Searching for activation functions.
\newblock Technical Report arXiv:1710.05941v1, Google Brain, 2018.

\bibitem{EgerEtAl:TimeToSwish:ProcEMNLP}
S.~Eger, P.~Youssef, and I.~Gurevych.
\newblock Is it time to swish? comparing deep learning activation functions
  across nlp tasks.
\newblock In {\em Proceedings of the 2018 Conference on Empirical Methods in
  Natural Language Processing (EMNLP), Brussels (Belgium)}, pages 4415--4424.
  Association for Computational Linguistics, 2018.

\bibitem{ChiengEtAl:FlattenTSwisshThresholdedSwish:JournAdvIntellSystems2018}
H.H. Chieng, N.~Wahid, O.~Pauline, and S.R.K. Perla.
\newblock Flatten-{T S}wish: a thresholded {ReLU-Swish}-like activation
  function for deep learning.
\newblock {\em International Journal of Advances in Intelligent Informatics},
  4(2):76--86, 2018.

\bibitem{Schneider2009_MatrixLearning}
P.~Schneider, B.~Hammer, and M.~Biehl.
\newblock Adaptive relevance matrices in learning vector quantization.
\newblock {\em Neural Computation}, 21:3532--3561, 2009.

\bibitem{Villmann2002d}
B.~Hammer and T.~Villmann.
\newblock Generalized relevance learning vector quantization.
\newblock {\em Neural Networks}, 15(8-9):1059--1068, 2002.

\bibitem{Villmann2018j}
S.~Saralajew, L.~Holdijk, M.~Rees, M.~Kaden, and T.~Villmann.
\newblock Prototype-based neural network layers: Incorporating vector
  quantization.
\newblock {\em Machine Learning Reports}, 12(MLR-03-2018):1--17, 2018.
\newblock {ISSN:1865-3960, http://www.techfak.uni-bielefeld.de/ $\tilde{ }$
  fschleif/mlr/mlr$\_$03$\_$2018.pdf}.

\bibitem{Villmann2019c}
T.~Villmann, J.~Ravichandran, A.~Villmann, D.~Nebel, and M.~Kaden.
\newblock Investigation of activation functions for {Generalized Learning
  Vector Quantization}.
\newblock In A.~Vellido, editor, {\em Proceedings of the 13th International
  Workshop on Self-Organizing Maps and Learning Vector Quantization, Clustering
  and Data Visualization, WSOM+2019, Barcelona}, page submitted. Springer
  Berlin-Heidelberg, 2019.

\bibitem{ElfwingEtAl:SwishforNeuralNetworksFunctionApproximation:NeuralNetworks2018}
S.~Elfwing, E.~Uchibe, and K.~Doya.
\newblock Sigmoid-weighted linear units for neural network function
  approximation in reinforcement learning.
\newblock {\em Neural Networks}, 107:3--11, 2018.

\bibitem{ZhangEtal:EfficientNeuralNetworkRobustnessCertificationWithGeneralActivationFunctions:NIPS2018_7742}
H.~Zhang, T.-W. Weng, P.-Y. Chen, C.-J. Hsieh, and L.~Daniel.
\newblock Efficient neural network robustness certification with general
  activation functions.
\newblock In S.~Bengio, H.~Wallach, H.~Larochelle, K.~Grauman, N.~Cesa-Bianchi,
  and R.~Garnett, editors, {\em Advances in Neural Information Processing
  Systems 31}, pages 4944--4953. Curran Associates, Inc., 2018.

\bibitem{CookSoftmax2011}
J.~Cook.
\newblock Basic properties of the soft maximum.
\newblock Working Paper Series~70, UT MD Anderson Cancer Center Department of
  Biostatistics, 2011.
\newblock http://biostats.bepress.com/mdandersonbiostat/paper70.

\bibitem{Villmann2013m}
M.~Lange and T.~Villmann.
\newblock Derivatives of $l_p$-norms and their approximations.
\newblock {\em Machine Learning Reports}, 7(MLR-04-2013):43--59, 2013.
\newblock {ISSN:1865-3960, http://www.techfak.uni-bielefeld.de/ $\tilde{ }$
  fschleif/mlr/mlr$\_$04$\_$2013.pdf}.

\bibitem{MaasEtAl:RectifierNonlinearitiesImproveNeuralNetworkAcousticModels:ICML2013}
A.L. Maas, A.Y. Hannun, and A.Y. Ng.
\newblock Rectifier nonlinearities improve neural network acoustic models.
\newblock In {\em Proc. ICML-Workshop for on Deep Learning for Audio, Speech,
  and Language Processing}, volume~28 of {\em Proceedings of Machine Learning
  Research}, 2013.

\bibitem{KrierFeatureSelectionICA2008}
C.~Krier, F.~Rossi, D.~François, and M.~Verleysen.
\newblock A data-driven functional projection approach for the selection of
  feature ranges in spectra with ica or cluster analysis.
\newblock {\em Chemometrics and Intelligent Laboratory Systems}, 91(1):43--53,
  2008.

\bibitem{Landgrebe2003a}
D.A. Landgrebe.
\newblock {\em Signal Theory Methods in Multispectral Remote Sensing}.
\newblock Wiley, Hoboken, New Jersey, 2003.

\bibitem{UCIRepository}
A.~Asuncion and D.J. Newman.
\newblock Uc irvine machine learning repository.
\newblock http://archive.ics.uci.edu/ml/.

\end{thebibliography}

\end{document}